\documentclass[10pt,twocolumn,letterpaper]{article}

\usepackage{cvpr}
\usepackage{times}
\usepackage{epsfig}
\usepackage{graphicx}
\usepackage{amsmath}
\usepackage{amssymb}
\usepackage{algorithm}
\usepackage{algorithmic}
\usepackage{ctable}
\usepackage{caption}
\usepackage{array}
\usepackage{flushend}
\usepackage{multirow}
\usepackage{xcolor}
\usepackage{multirow}
\usepackage{authblk}

\usepackage[pagebackref=true,breaklinks=true,letterpaper=true,colorlinks,bookmarks=false]{hyperref}

\DeclareMathOperator*{\argmin}{arg\,min}

\cvprfinalcopy 

\def\cvprPaperID{2312} 

\ifcvprfinal\fi
\begin{document}
	
\newcommand{\alla}[1]{\textcolor{blue}{#1}}
\newcommand{\ALLA}[1]{\textcolor{blue}{\textbf{Alla: #1}}}
\newcommand{\leonid}[1]{\textcolor{green}{#1}}
\newcommand{\HR}[1]{\textcolor{cyan}{\textbf{Helge: #1}}}
\newcommand{\hr}[1]{\textcolor{cyan}{#1}}
\newcommand{\todo}[1]{\textcolor{red}{#1}}
\newcommand{\TODO}[1]{\textcolor{red}{\bf TODO: #1}}

\newcommand{\comment}[1]{}
\newcommand{\parag}[1]{\vspace{-3mm}\paragraph{#1}}

\title{Front2Back: Single View 3D Shape Reconstruction via Front to Back Prediction}

\newcommand*{\affaddr}[1]{#1} 
\newcommand*{\affmark}[1][*]{\textsuperscript{#1}}
\newcommand*{\email}[1]{\texttt{#1}}

\author{%
	Yuan Yao\affmark[1]
	Nico Schertler\affmark[1]
	Enrique Rosales\affmark[1, 2]
	Helge Rhodin\affmark[1]
	Leonid Sigal\affmark[1, 3]
	Alla Sheffer\affmark[1]
	\\
	\affaddr{\affmark[1]University of British Columbia}\\
	\affaddr{\affmark[2]Universidad Panamericana}\\
	\affaddr{\affmark[3]Canada CIFAR AI Chair, Vector Institute}\\
	\email{\{rozentil, nschertl, albertr, rhodin, lsigal, sheffa\}@cs.ubc.ca}\\

}
%
\maketitle


\begin{abstract}

Reconstruction of a 3D shape  from a single 2D image is a classical computer vision problem, whose difficulty stems from the inherent ambiguity of recovering occluded or only partially observed surfaces. 
Recent methods address this challenge through the use of largely unstructured neural networks that effectively distill conditional mapping and priors over 3D shape. 


In this work, we induce structure and geometric constraints by leveraging three core observations:
(1) 
the surface of most everyday objects is often almost entirely exposed from pairs of typical opposite 
views; (2) everyday 
objects often exhibit global reflective symmetries which can be accurately predicted from single views; 
(3) opposite orthographic views of a 3D shape share consistent silhouettes.


Following these observations, we first predict
orthographic 2.5D {\em visible surface} maps (depth, normal and silhouette) from perspective 2D images, and detect global reflective symmetries in this data;
second, we predict the back facing depth and normal maps using as input the front maps and, when available, the symmetric reflections of these maps;
and finally, we reconstruct a 3D mesh from the union of these maps using a surface reconstruction method best suited for this data.   


Our experiments demonstrate that our framework outperforms state-of-the art approaches for 3D shape reconstructions from 2D and 2.5D data in terms of input fidelity and details preservation.
Specifically, we achieve 12\% better performance on average in ShapeNet benchmark dataset \cite{chang2015shapenet}, and up to 19\% for certain classes of objects (\eg,  chairs and vessels).


\end{abstract}


\begin{figure}[t]
    \centering
    \includegraphics[width=\linewidth]{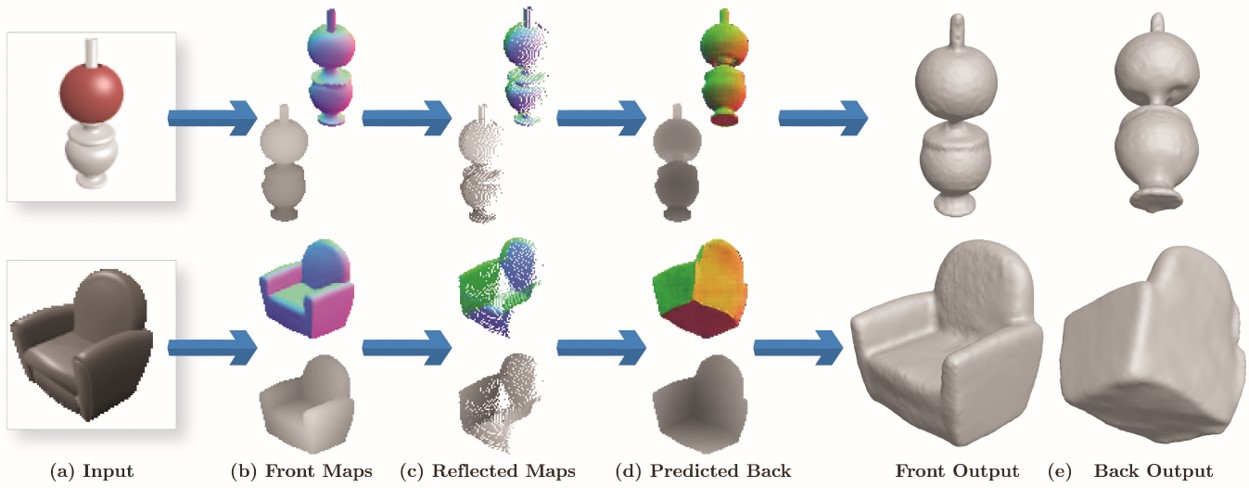}
    \caption{{\bf Front2Back} reconstructs 3D surfaces (left) from single 2D images (a) by first computing the corresponding front facing 2.5D normal, depth and silhouette (not shown) visible surface maps (b) and their symmetric reflections (when present) (c) ; it then predicts the back (opposite) view maps (d); and finally fuses all these intermediate maps via surface reconstruction to produce a watertigth 3D surface (e). 
    Our approach is able to recover more detailed and faithful 3D shape reconstructions compared to state-of-the-art.  
       } 
    \label{fig:teaser}
    \label{fig:0}
\end{figure}

\section{Introduction}
\label{sec:intro}

Humans are amazingly adept at predicting the shape of 3D objects from a single 2D image, a task made particularly challenging by the need to predict occluded surface geometry. 
Early computer vision researchers hypothesized that humans achieve this goal by employing a series of intermediate representations, of progressively increasing complexity, and envision the 2.5D {\em visible surface} geometry as one of these representations \cite{marr1982vision}.
This argument motivates multi-step 3D reconstruction pipelines that use 2.5D visible surface as an intermediate representation \cite{zhang2018shaphehd}. Our novel {\em Front2Back} algorithm targets complete 3D surface reconstruction from such an intermediate representation, 
encoded via multiple depth and normal maps that capture global reflective symmetries (Figure~\ref{fig:teaser}bc). We create a complete image to 3D reconstruction framework by combining Front2Back with a data-driven predictor that computes accurate 2.5D visible surface geometry from raw 2D images  undoing perspective effects (Figure~\ref{fig:teaser}a). 

Recent methods directly learn 3D shape representation from raw 2D images \cite{chen_cvpr19,choy20163d,groueix2018,Mescheder19,tulsiani2017multi,NIPS2016_6096} 
or intermediate 2.5D visible surface maps~\cite{sun2018pix3d,wu2017marrnet,wu2018learning,zhang2018shaphehd} and represent 3D shapes using voxelizations or implicit functions (defined over $\mathbb{R}^3$). Such models often implicitly distill shape prior information into the network through a large set of 2D-3D paired exemplars obtained from ShapeNet \cite{chang2015shapenet} or similar large scale 3D model repository.
%
These methods, collectively, produced an increasingly impressive set of results on single-view 3D shape reconstruction, but are limited in terms of fidelity, a limitation which can be largely attributed to the resolution of the resulting voxel grid, as well as to the inherent complexity of learning volumeteric ($\mathbb{R}^3$) information directly from images ($\mathbb{R}^2$). 
While the dimensionality challenge can be sidestepped via the use of  2D (texture) atlases describing the output point-clouds ~\cite{groueix2018} or the use of deformable templates \cite{wang2018pixel2mesh,wang20193dn}, these approaches face other challenges.
Notably, none of the existing methods take explicit advantage of geometric cues correlating visible and occluded surface parts; nor leverage strong perceptual cues in the input image, such as symmetry, which has been shown to be integral for the human 3D perception~\cite{Pizlo2014}.

Similar to ~\cite{groueix2018} we utilize an image-centric 2.5D intermediate data representation which allows us to describe output point-clouds precisely.
However, contrary to all the above approaches we explicitly leverage geometric and perceptual linkages between visible and occluded surface geometry.
The core observation behind our framework is that most
everyday objects can be {\em almost} entirely described by pairs of oppositely oriented height-fields; moreover this property holds for most orientation choices\footnote{In our experiments, a random sample of 520 shapes from ShapeNet core had on average 80\% of their surface visible from 95\% of random opposite view pairs.}. 
Therefore, a pair of 2.5D visible surface images taken from opposite views frequently describes a (nearly) complete 3D model (Figure~\ref{fig:teaser}b,~\ref{fig:teaser}c). 
Notably for orthographic projection, the silhouettes of such opposite images always align. 
Consequently, the problem of reconstructing 3D models from single, non-accidental,  2.5 orthographic visible surface images can for many classes of objects be effectively reduced to the {\em Front2Back} problem:  correctly recovering the back facing 2.5D visible surface given a front facing one. 
The core advantage of this approach is that instead of directly obtaining 3D data from a 2D or a 2.5D input, a process that is still not well understood, we can now recast our core problem as synthesizing one type of, silhouette aligned, image (back) from another (front), a task that had been successfully tackled by recent image-to-image translation networks.
%
Lastly, and critically, many everyday shapes exhibit global reflective symmetries, which can be accurately detected from front visible surface maps alone. 
These symmetries allow viewers to predict backfacing geometry as well as to complete details not available in either front or back views. 

\comment{
\hr{
Our contributions to realizing Front2Back are four-fold:
1) A novel framework that combines recent learning-based approaches and more traditional geometric processing for symmetry detection and surface reconstruction from points plus normals. 
2) We introduce an intermediate representation as the orthographic 2.5D {\em back} view of the object, which we obtain using a state-of-the-art image-to-image translation architecture.
3) In both back view prediction and final surface reconstruction, we leverage symmetries detected on the front view to significantly improve the performance.
4) We introduce a predictor that converts 2D images taken with a regular, perspective camera into
2.5D orthographic surfaces,
which lets us apply Front2Back on natural images.}

Front2Back leads to significant improvements in the resulting 3D shape, compared to competing state-of-the-art reconstruction methods, 
\alla{here goes our numbers brag.}}

\comment{
\vspace{0.1in}
\noindent
{\bf Contribution:} 
\leonid{
Our contributions are four-fold:
(1) We propose a novel framework that combines recent learning-based approaches and more traditional geometric processing for symmetry detection and surface reconstruction from points plus normals. 
(2) We introduce an effective intermediate representation consisting of the orthographic 2.5D {\em back} view of the object, which we obtain using a state-of-the-art image-to-image translation architecture.
(3) In both, back view prediction and final surface reconstruction, we leverage symmetries detected on the front view to significantly improve the performance.
(4) We introduce a network that takes 2D images taken with a regular, perspective, camera and predicts corresponding
2.5D orthographic front surfaces,
which allows us to apply our method (Front2Back) on natural RGB images.}
}

Following these observations, Front2Back employs an image-to-image translation network to predict back view 2.5D visible surface maps (depth+normal) from the input front 2.5D representation. 
It then reconstructs a 3D surface mesh fusing together the front and back geometry information using a reconstruction framework capable of 
producing watertight surfaces by combining positional and normal information 
~\cite{kazhdan2013screened}. 
In both the prediction of the intermediate 2.5D back view maps and the reconstruction itself, it leverages global reflective symmetries, algorithmically detected 
on the front visible surfaces, when present: reflected front surface maps (Figure~\ref{fig:teaser}) are provided as auxiliary input for back prediction, and used as an additional source of data for subsequent reconstruction. 
To achieve direct 3D reconstruction from raw images we combine our method with a learning based framework that for each input image computes the corresponding orthographic 2.5D visible surfaces rectifying perspective distortion present in these images. 
Our approach leads to significant improvements in the resulting 3D shape reconstruction, compared to competing state-of-the-art methods, 
resulting in 12\% improvements on average, measured using mesh-to-mesh distance~\cite{Cignoni1998} and up to 19\% improvements on certain ShapeNet object classes, compared to the closest competitors.

\noindent
{\bf Contribution:} 
Our core contribution is a novel framework for 2.5D visible surface map to 3D reconstruction, which combines recent learning-based approaches with more traditional geometric processing and produces results superior to the state-of-the-art. 
This contribution is made possible by two core technical innovations: (1) novel and effective intermediate representation consisting of the orthographic 2.5D {\em back} view of the object, which we obtain using a state-of-the-art 
image-to-image translation architecture; and (2) utilization of symmetries detected in the front view to significantly improve the performance of both the 2.5D {\em back} view prediction and overall surface reconstruction. 

\comment{
Our core contribution is a novel framework for 2.5D visible surface map to 3D reconstruction, which combines recent learning-based approaches with more traditional geometric processing and produces results superior to the state-of-the-art. This contribution is made possible by the use of two key intermediate components: predicted back view and reflected symmetric maps of the target object, robustly and successfully generated from the  front facing maps.  
}

\comment{
\vspace{0.1in}
\noindent
{\bf Contribution:} 
Our contribution is three-fold. First, we introduce a novel framework for 2.5D visible surface map to 3D reconstruction, which, at its core, is a combination of recent learning-based approaches and more traditional geometric processing (symmetry detection and surface reconstruction from points plus normals). 
This framework has several unique advantages, including the ability to produce more accurate and detailed reconstructions compared to the art alternatives. Second, in doing so, we introduce an intermediate representation of 2.5D {\em back} view of the object, which we obtain using state-of-the-art GAN-based image-to-image translation method. The concept of using a back, or opposite, view prediction as an intermediate step for complete 3D reconstruction from a single image is, to our knowledge, novel. We highlight that opposing views, combined, frequently describe most, if not all, of a target object's surface. 
Third, in both  back view prediction and final surface reconstruction, we leverage symmetries detected on front view maps to significantly improve the performance. Symmetry is an important perceptual cue, that, to the best of our knowledge, has not been utilized in neural-based shape reconstruction. 
\alla{mention resolution advantage and maybe multi-class training if applies}
}


\section{Related Work}


We build upon a large body of work on single-view 3D surface reconstruction, view synthesis, and image-to-image translation. 

\vspace{0.07in}
\noindent
{\bf Single-view 3D Surface Reconstruction.}
%
Reconstructing 3D models from single view 2D images, or 2.5D depth (and normal) maps, is a difficult and ill-posed problem. Recent learning based methods show promising results in addressing this challenge~\cite{what3d_cvpr19}.  
While many methods predict 3D shape representations directly from 2D images, \eg, ~\cite{chen_cvpr19,choy20163d,FanSG17,Girdhar16b,groueix2018,Mescheder19,tulsiani2017multi,wang2018pixel2mesh,NIPS2016_6096,NIPS2016_6206},
others, \eg,~\cite{sun2018pix3d,wu2017marrnet,wu2018learning}  first reconstruct the 2.5D visible surface (typically representing it via depth and normal maps) and then use this intermediate representation as a stepping stone toward complete 3D reconstruction. 
Many methods in both categories represent the reconstructed shapes using voxels, \eg,~\cite{Girdhar16b,wu2017marrnet,wu2018learning,NIPS2016_6206} or limited-depth octrees \cite{Wang-2018-AOCNN}.
The accuracy of the reconstructions produced by these methods is limited by the finite resolution of the voxel or octree cells, limiting the methods' ability to capture fine details.
Template based methods~\cite{sinha2017surfnet,wang2018pixel2mesh,wang20193dn} perform well when the topology of the input template matches that of the depicted shape, but are less suited for cases where the target topology is not known {\em a priori}. 
Recent implicit surface based methods ~\cite{chen_cvpr19,Mescheder19} strive  for resolution independence  but require watertight meshes for training. Since the vast majority of meshes in the wild are far from watertight\footnote{Less than 2.5\% of the 7K+ models in the test split of ShapeNet core 
 are watertight.} instead of training directly on this data they use watertight approximations which 
necessarily deviate from the originals. This deviation can potentially bias surface prediction.   
Atlas based \cite{groueix2018} reconstruction avoids these pitfalls, but exhibits similar accuracy levels. \comment{ (Section~\ref{sec:results}).}  Point or depth-only maps based methods  \cite{FanSG17,shin2018pixels,lin2018learning} produce collections of points close to the target objects surface; however surface reconstruction from unoriented points is a challenging problem in  its own right~\cite{berger2014state}, thus when attempting to recover the surface from cloud the output quality decreases dramatically \cite{shin2018pixels}.
Our method uses depth plus normal maps as intermediate representation, works well on shapes with diverse topology, has no bounds on accuracy beyond the input image resolution; and can be directly trained on models with arbitrary connectivity and non-manifold artifacts, avoiding the bias introduced by implicit function based approaches. 
\comment{Most importantly, our comparisons show that it produces more accurate reconstructions than state-of-the-art methods (Figure~\ref{fig:compare}, Section~\ref{sec:results}).}

Two recent methods reconstruct front and back depth map images of human subjects from photographs \cite{gabeur2019,natsume_siclope:_2019}. Our method differs from those in utilizing a normal map alongside depth, symmetric cues, and normalized, othographic coordinates obtained through perspective correction to predict depth maps across a wide range of classes/views.

Symmetries are used in \cite{wang20193dn}, but only for regularizing the output at training time, unlike our explicit integration of symmetry as a cue for reconstruction, which requires symmetry detection on the input image at test time.


%


\begin{figure*}[th]
  \centering
  \includegraphics[width=1.0\textwidth]{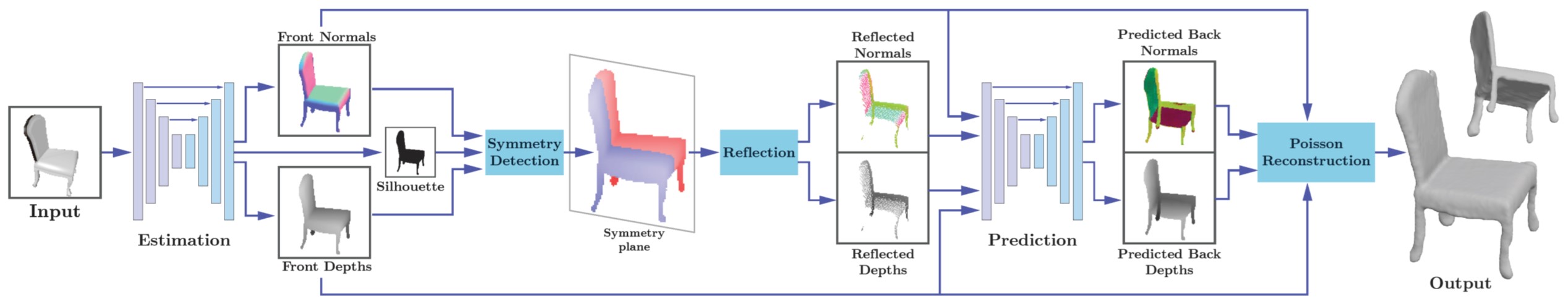}
  \caption{{\bf Algorithm Stages:} (left to right): input perspective image; 2.5D orthographic map (depth+normal+silhouette) prediction; detection of reflective symmetries (regions on the left and right of the plane colored blue and red respectively); back map prediction; final surface reconstruction.}
  \label{fig:methodOverview}
\end{figure*}

\vspace{0.07in}
\noindent
{\bf View Synthesis and Shape Completion.}
Our core task of predicting back view 2.5D surface can be seen as a special case of alternative view synthesis~\cite{dosovitskiy2015learning, eigen2014depth, greene1986environment, liu2016learning, zhou2016view}.  Most such methods aim to predict views that have similar viewpoints to the original. Recent approaches, \cite{chen2019monocular} render compelling images and depth maps from viewpoints that differ by up to $40^\circ$ from the original. Surface reconstruction from depth maps alone  \cite{lin2018learning,shin2018pixels} suffers from similar pitfalls as one from unoriented clouds \cite{berger2014state}.
In contrast to these settings we seek to recover the exact opposite view, where the only overlap between the observed surfaces is along the outer silhouette.
By combining these strategically selected views and utilizing symmetry we  successfully compute oriented point-clouds, allowing for the use of more robust reconstruction 
methods~\cite{kazhdan2013screened} and producing far superior reconstruction results.  

%
%
Learning based shape completion methods, \eg, \cite{han2017high,park2019deepsdf}, attempt to extend partial surfaces across poorly captured regions to produce a complete shape representation. These methods are designed to operate on 3D shape inputs, are typically limited to filling relatively small holes, and employ shape priors for ambiguity resolution. In contrast, our Front2Back step predicts a complete surface from only front facing depth+normal maps and uses an intermediate 2.5D back map to achieve this goal.

\parag{Image-to-image translation.} 
Image-to-image translation  \cite{huang2018multimodal, isola2017image, lee2018diverse, wang2017high, zhu2017unpaired} is a powerful tool for synthesizing new images from existing ones for applications such as image-from-sketch synthesis~\cite{sangkloy2017scribbler} and make-up application~\cite{chang2018pairedcyclegan}. While typical translation methods aim to preserve the view and the content of the original image and only change 
some of their visual attributes, generating back views from front ones requires significant changes in the depth and normal content, a much more challenging task.

\section{Approach}
\label{sec:method}

\comment{This section gives a high-level overview of our approach.}
\comment{Front2Back}
Our method takes as input a single perspective 2D image and generates a 3D mesh of the corresponding object, using four key steps (Figure~\ref{fig:methodOverview}).
We start the process by predicting orthographic silhouette, depth, and normal maps of the portion of the target object's surface visible in the input image (Section \ref{sec:img2front}). 
We proceed to locate a global 3D reflective symmetry plane from this visible surface, if one exists (Section~\ref{sec:symmetry}).  We use the located plane (if detected) to
infer the occluded parts of the shape whose symmetric counterparts are visible, by reflecting the input maps over the symmetry plane to get a second set of depth and normal maps (hereon referred to as reflected maps).
We mask all depth and normal maps using the silhuette , denoting all pixels outside as background, and use these maps an input to our core back prediction stage.
The prediction stage takes this input 
and generates new depth and normal maps for the back view, the exact opposite of the input front view (Section~\ref{sec:back}).
We perform this prediction using a variant of conditional generative adversarial networks for image to image translation.
Finally, we combine the front view maps, reflected maps, and predicted back view maps to extract the corresponding oriented point cloud and reconstruct the surface from this cloud (see Section~\ref{sec:surface} for details).

\subsection{Orthographic Front View Prediction}
\label{sec:img2front}
For the 2D to 2.5D step, we adopt and train the 2.5D estimation network of \cite{wu2018learning}, using example input-output pairs of perspective images and corresponding same view direction orthographic depth, normal, and silhouette maps. Perspective rectification simplifies subsequent symmetry estimation and allows us to enforce the same silhouette constraint across all computed maps. 

We define the loss function as the sum of the three individual $L_1$ losses of the outputs. 
The original network is designed for noisy, real-life images and purposefully adds 
noise to input data to mimic real-life artifacts; since similar to most recent single view reconstruction 
methods our training image set is synthetic, for a fair comparison we disabled this feature.


\subsection{Symmetry Detection}
\label{sec:symmetry}

Reflective symmetry is a frequent feature of both organic and man made shapes. It plays a vital role in human perception - absent information to the contrary,  human observers expect symmetries observed in the visible parts of a surface to extend into occluded 
parts facilitating mental reconstruction of these occluded surfaces \cite{Hoffman2000}.  
Our method mimics this behavior by explicitly formulating a surface hypothesis derived from reflective symmetries detected on the front view maps.
Efficient, robust, and accurate detection of reflective symmetry planes from partial surfaces is a challenging geometry processing problem~\cite{Aiger:2008,mitra2006partial}. While existing methods are designed for dense point-clouds, we seek to detect symmetries on pixelated and thus heavily quantized data, which frequently has very low local resolution (\eg chair legs which are less than 5 pixels wide). Most critically, we seek to avoid false positives, as inaccurate reflected maps can severely impact our subsequent back prediction and reconstruction steps.

\begin{figure}
	\centering
	\includegraphics[width=\linewidth]{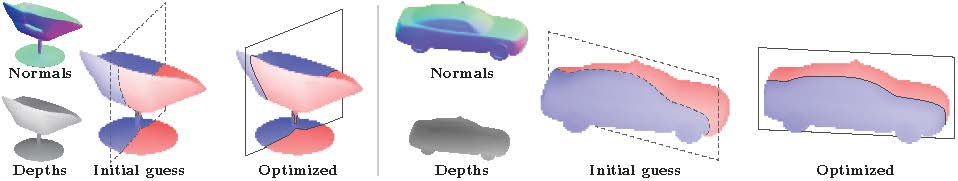}
	\caption{{\bf Symmetry plane detection.} Given front view maps (insets, left) we use plane-space clustering to obtain initial reflective planes (center, dashed) and obtain the final planes (solid,right) by using ICP iterations. 
	(For many inputs the improvement is more subtle.)}
	\label{fig:symmetryDetection}
\end{figure}

We design a targeted two-step reflective symmetry plane detection method, that addresses these challenges by combining information across all three front maps. 
We first estimate an approximate symmetry plane using a clustering-based approach, sped up using a variant of RANSAC. We refer to this plane as \textit{initial} plane. We then optimize this initial plane to better align the original oriented points with their reflected counterparts using an iterated closest point method (Figure \ref{fig:symmetryDetection}). We avoid false positives in both steps by usilizing two key constraints. We note that the silhouette map defines the visual hull of the target object, thus we expect any surface parts symmetric to parts of the visible surface to lie inside it; we thus filter out reflection planes that produce maps violating this constraint. Since our front maps are expected to be at least as accurate as the reflected ones, we similarly only consider planes that produce reflected maps that do not occlude the front surfaces (\ie have no points closer to the viewer at the same x-y locations), we refer to this constraint as visibility. 
\comment{ it imposes.}
We use the detected planes to generate reflected depth and normal maps. 

\vspace{0.1in}
\noindent
{\bf Initial plane.}
We use a clustering-based symmetry detection algorithm, inspired by~\cite{mitra2006partial} for estimating the initial symmetry plane. We first transform the front-facing normal and depth map into an {\em oriented} point set $P$. 
For each pair of oriented points $(\mathbf{p}_i, \mathbf{p}_j) \in P \times P$, we calculate the plane that best reflects $\mathbf{p}_i$ to $\mathbf{p}_j$. 
We then repeatedly sample subsets of such potential symmetry planes and cluster each subset using mean-shift clustering. 
We compute a Voronoi diagram of all planes with respect to the centers of the resulting clusters, and define the score of each center as the number of planes in its' respective cell. 
To select the initial plane among cluster centers obtained across all 
iterations we first discard all centers that produce reflected maps significantly violating visual hull or visibility constraints.
We then select as initial plane the center with the highest score. 
 
\vspace{0.1in}
\noindent
{\bf Optimization.}
Our initial plane is based on a sampling approach and only considers planes defined directly by point pairs, thus while close to the optimal, it can often be further improved.
We optimize the plane using a variant of classical ICP. At each iteration, for each point $\mathbf{p}\in P$, we first calculate  the point $\mathbf{r}_p \in P$ that is closest to its reflection
around the current symmetry plane $s$.
We prune all correspondences whose distance exceeds a user-definable threshold or whose normals are not reflected within a tolerance, and use the remaining correspondences $P_c \subseteq P$ to optimize for a symmetry plane that maps the points closer to their reflections, \ie,
\begin{equation}
    s' = \argmin_s \sum_{\mathbf{p} \in P_c} \left\lVert \rho_s(\mathbf{p}) - \mathbf{r}_p \right\rVert^2.
\end{equation}
We solve this minimization problem using gradient descent with back tracking line search to determine the step size and update the estimate for the symmetry plane with $s'$. We repeat the process until convergence (Figure \ref{fig:symmetryDetection}, right).

\vspace{0.1in}
\noindent
{\bf Determining Model Symmetry.}
 We classify the input as asymmetric if the resulting reflected maps violate visual hull or visibility constraints with tight thresholds, or generates a map which covers less than 40\% of the silhouette interior.
%
False symmetry can dramatically impact reconstruction quality, far more so than reconstruction with no symmetry information, motivating us to use fairly strict thresholds.
In practice, we classify about $\approx$ 20\% of the data as asymmetric. 

\subsection{Back View Maps Prediction}
\label{sec:back}

\begin{figure*}[ht]
	\centering
	\includegraphics[width=0.7\linewidth]{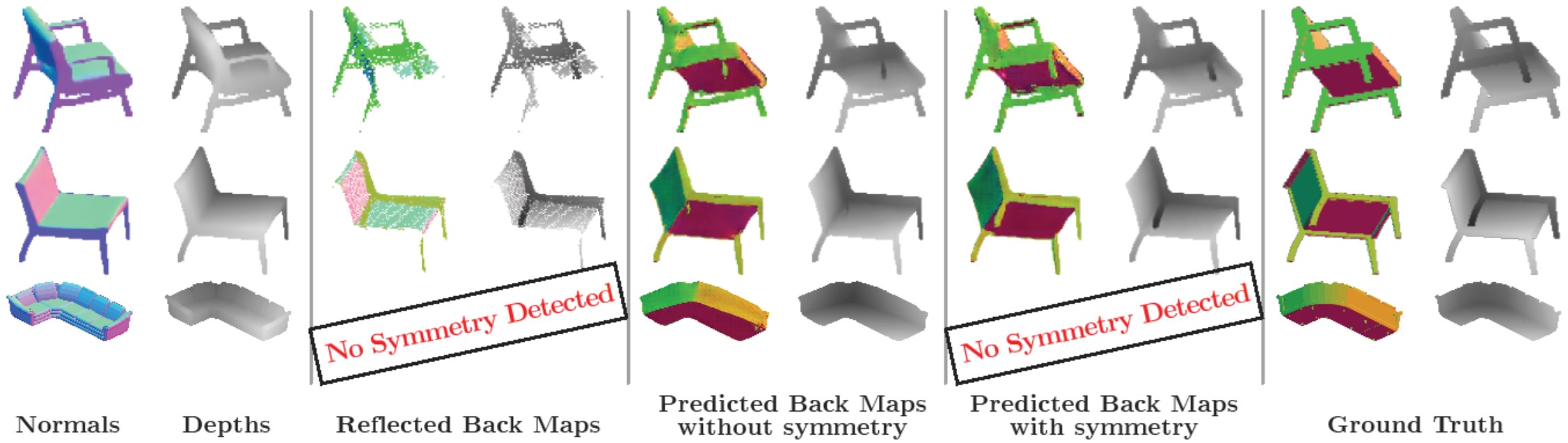}
	\caption{{\bf Back map prediction.}  Given front and reflected depth+normal maps our image-to-image translation network predict the corresponding back view maps; we correctly identify asymmetric models and predict the correspondig back maps usig a separately trained predictor which used only front maps as input.
	}
	\label{fig:intermediate results}
\end{figure*}

Recent deep learning-based image-to-image translation methods focus on the translation between different domains.  \comment{ using a static view. }
In our method, we demonstrate that such deep neural networks can also be used to learn the mapping between 2.5D representations in opposing views. 
\comment{images of the same domain from strictly opposite views.}

Our learning model is similar to~\cite{isola2017image}, which is based on a conditional generative adversarial network (cGAN). The architecture consists of two networks: a generator and a discriminator. The discriminator is trained on a training data set to classify an input image into either real or fake w.r.t. to training data set. The generator is trained to produce images that the discriminator evaluates as real. 
Our goal is to predict the back view normal and depth map from front view maps. To integrate symmetry information, we also feed the reflected depth and normal maps into the network when symmetries are detected (see end of Section~\ref{sec:symmetry}).
Hence the input to the generator is a depth-$4$ or -$8$ image, depending on symmetry circumstance, and the output is a depth-$4$ image encoding predicted back depth and normal maps. To train the network, we use a loss function $L$ that comprises terms for each of the individual aspects of our prediction problem:
\begin{multline}
L = w_{GAN}L_{GAN}+w_dL_d + w_nL_n. 
\label{eq:gan}
\end{multline}
The GAN loss $L_{GAN}$ is the traditional loss used for generative adversarial networks that steers the interplay between generator and discriminator.
The two similarity loss functions for depth $L_d$ and normals $L_n$ aim to measure the pixel-wise differences between predicted and ground truth maps. 
In our experiments, we set $w_{GAN} = 1$, $w_n = 100$, and $w_d = 1000$. 
In the following, we present the loss functions in more detail.

\vspace{.1in}
\noindent
{\bf Adversarial loss.}
We use the traditional adversarial loss as presented in \cite{goodfellow2014generative}. Given the front view normal and depth maps $\mathbf{N}_f$ and $\mathbf{D}_f$, the reflected back view maps $\mathbf{N'}_b$ and $\mathbf{D'}_b$, we define the adversarial loss as:
\begin{equation}
\begin{split}
L&_{GAN}(G,D) = \\
&\mathbb{E}_{(\mathbf{N}_f, \mathbf{D}_f)}\left[\log(D(\mathbf{N}_b, \mathbf{D}_b))\right] +\\
&\mathbb{E}_{(\mathbf{N}_f, \mathbf{D}_f, \mathbf{N'}_b, \mathbf{D'}_b)}\left[\log(1-D(G(\mathbf{N}_f, \mathbf{D}_f, \mathbf{N'}_b, \mathbf{D'}_b)))\right]
\end{split}
\end{equation}

\noindent
{\bf Similarity loss.}
Similar to many existing image-to-image translation tasks, we use L1 loss between the output and ground truth images as a similarity measure for the depth maps. Since the normals represent an orientation, for which deviation is more accurately represented by an angle, we use cosine similarity for the normal maps. Given the predicted back view maps $\mathbf{\hat{N}}_b$ and $\mathbf{\hat{D}}_b$ produced by the generator and the ground truth normal and depth images $\mathbf{N}_b$ and $\mathbf{D}_b$:
\begin{equation}
\begin{split}
L_{d} &= \left\lVert \mathbf{D}_b - \mathbf{\hat{D}}_b \right\rVert_1  \\
L_{n} &= \left[\mathbf{N}_b, \mathbf{\hat{N}}_b\right]_\mathrm{cos}, \\
\end{split}
\end{equation}
where
\begin{equation}
\begin{split}
[\mathbf{A}, \mathbf{B}]_\mathrm{cos} &= \sum_{i,j} { \left(1 - \frac{\mathbf{A}(i,j) \cdot \mathbf{B}(i,j)}{\left\lVert \mathbf{A}(i,j) \right\rVert \cdot \left\lVert \mathbf{B}(i,j) \right\rVert} \right).}
\end{split}
\end{equation}

\comment{
\alla{do we use this?}
\vspace{0.1in}
\noindent
{\bf Consistency loss.}
To maintain consistency between the output normal and depth maps, we add a consistency loss. This loss penalizes tangents of the predicted depth map that are not orthogonal to the predicted normals.
\begin{equation}
\begin{split}
L_{c} = & \sum_{i,j} \mathbf{\hat{N}}_b(i,j) \cdot \\
        & \hspace{-3mm}  \left(w\left(\frac{\partial \mathbf{\hat{D}}_b}{\partial x} (i,j)\right) T_x(i,j) + w\left(\frac{\partial \mathbf{\hat{D}}_b}{\partial y} (i,j)\right) T_y(i,j) \right), \\
\end{split}
\end{equation}   
where
\begin{equation}
w(t) =  \frac{1}{\sigma \sqrt{2 \pi}}\exp\left(\frac{-t^2}{2 \sigma}\right)
\end{equation}

The term $T_{x/y}(i,j)$ represents the 3D tangents reconstructed from the predicted depth image in the $x$ and $y$ direction at pixel $i, j$. The Gaussian weight $w$ is added to reduce the effect of the consistency loss in non-smooth regions, where estimating normals from depth differences becomes unreliable.}

\subsection{Surface Reconstruction}
\label{sec:surface}

We fuse the per-pixel positions and normals form the front, reflected, and predicted back maps to generate an oriented point cloud and use screened Poisson surface reconstruction~\cite{kazhdan2013screened} to surface this input. 
To produce closed meshes we use Dirichlet boundary conditions, and use an interpolation weigth of $4$ to promote interpolation of input points. 
Our fusion process automatically corrects quantization artifacts and inaccuracies in the computed maps that can lead to catastrophic reconstruction failures, and accounts for point density assumptions made by typical reconstruction methods.

\vspace{0.07in}
\noindent
{\bf Fusion.}
Similar to many other methods, screened Poisson reconstruction~\cite{kazhdan2013screened} expects input points originating from the same surface to be closer to one another than to points originating from an oppositely oriented surface. To satisfy this property given any pair of map points with oppositely oriented normals (one front facing and one back facing) which are within image-space distance of less than two pixels of one another, we ensure that the depth difference between them is at least two pixels, by moving the back facing point backward if needed. 
Following the visibility prior, we expect the front map points to be more reliable and closer to the viewer than points from other sources, thus we discard reflected and back map points its they are closer to the viewer than front map points at the same $x,y$. Following the same argument, we expect back map points to be farthest from the viewer, however in cases of depth conflicts we trust reflected maps more that back prediction. Thus we discard back map points that are closer to the viewer than reflected map points at the same $x,y$.
Lastly, we remove outliers  which we classify as points which are above a fixed threshold away from all four of their immediate image-space neighbors in the same map along the depth axis. The logic behind this criterion is that while we expect the depth maps to exhibit discontinuities, we do expect local features to be larger than one pixel. 

More fine grain implementation details are provided in the supplementary material.

\section{Experiments}

We tested our methods on a large body of models across different classes
 and performed both qualitative and quantitative comparisons, as well as ablation studies demonstrating the impact of the different algorithmic choices made. 
Additional results, including ones generated from $256\times256$ images,  are provided in the supplementary material. 

\noindent
{\bf Dataset.}  In our experiments we use the ShapeNet Core dataset~\cite{chang2015shapenet} and its official training/testing split, which includes 13 object categories. 
To compare against both methods that use this split and those that use the training/testing split of Choi \etal \cite{choy20163d} in the results below we report all quantities on the intersection of the two test sets that none of the models have seen in training.  
To be comparable, we train our orthographic front view prediction using the training split of the images corresponding to the training models provided by Choi \etal \cite{choy20163d} at resolution $137\times 137$ pixel. We follow \cite{Mescheder19} and others and use the first, random view, provided by Choi \etal. for the test set.
To generate both the front and back view maps for training, we rendered orthographic depth and normal map from these views and their opposites. We use the canonical symmetry plane  ($yz$)  of the training shapes that are symmetric around it for generating reflected maps for training.

\noindent
{\bf Metrics.} For evaluation we use the mesh-to-mesh symmetric $L_1$ distance (MD) ~\cite{Cignoni1998} and Chamfer $L_1$ distance (CD) between the ground truth and reconstructed meshes. We measure MD using Metro~\cite{Cignoni1998}, a well established measurement tool in the geometry processing community, using default Metro parameters. We measure CD  using the implementation provided by \cite{Mescheder19} across 100K evenly sampled points on the two meshes. The core difference between these metrics is that Metro looks for closest distance from a sample point on one mesh to any of the triangles on the other, while CD only considers sample to sample distances and thus is inherently more dependent on sampling density and quality. We report MD using the diagonal of the ground truth bounding box as unit 1, and follow the convention in \cite{Mescheder19} when reporting CD. 

\noindent 
{\bf Implementation Details.} 
We use `adam' optimizer with learning rate 0.0001 and batch size of 4 for training our orthographic front view predictor. 
We use the loss and corresponding architecture and parameter settings from \cite{wu2018learning}.
For symmetry detection we use 20 iterations each with 8K plane samples when computing initial plane up to 400 iterations for ICP; for
correspondence pruning we use thresholds of 4 pixels and $60^\circ$ on position and normals. We reject reflection planes as violating the visual hull threshold, if over 5\% of the reflected pixels are at least 5 pixels outside the silhouette,  and reject them as violating visibility if over $15\%$ of the pixels are in front of the front view.
We use `adam'  optimizer with learning rate 0.0002 and batch size of 1 for training our back prediction network, and use 5 random views out of the 24 provided by \cite{choy20163d}.
For final reconstruction step outlier removal we use threshold of 4 pixel sizes.
These parameters are kept constant for all experiments in the paper. 

\begin{figure}[h]
    \centering
    \includegraphics[width=.9\linewidth]{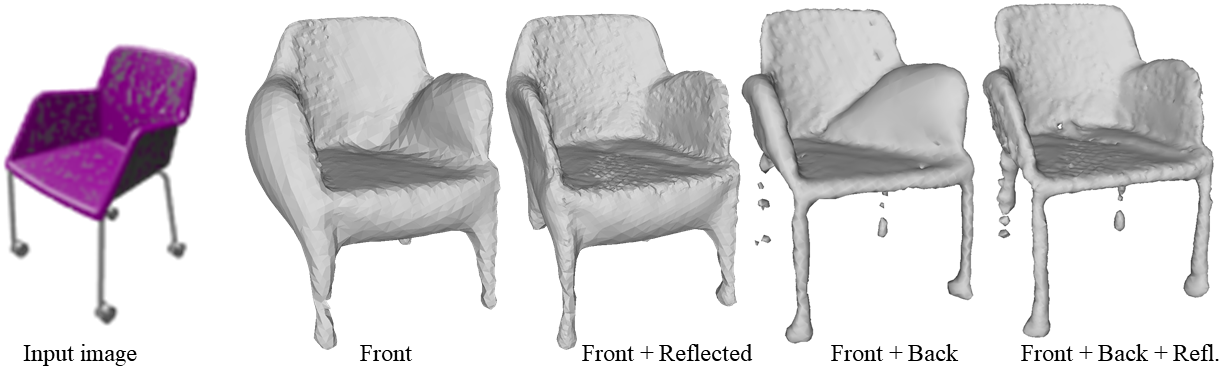}
   \caption{{\bf Final reconstruction ablation.} Each of the chosen intermediate maps improves the final surface reconstruction quality.}
    \label{fig:ablation}
\end{figure}

\begin{figure}[h]
    \centering
    \includegraphics[width=0.7\linewidth]{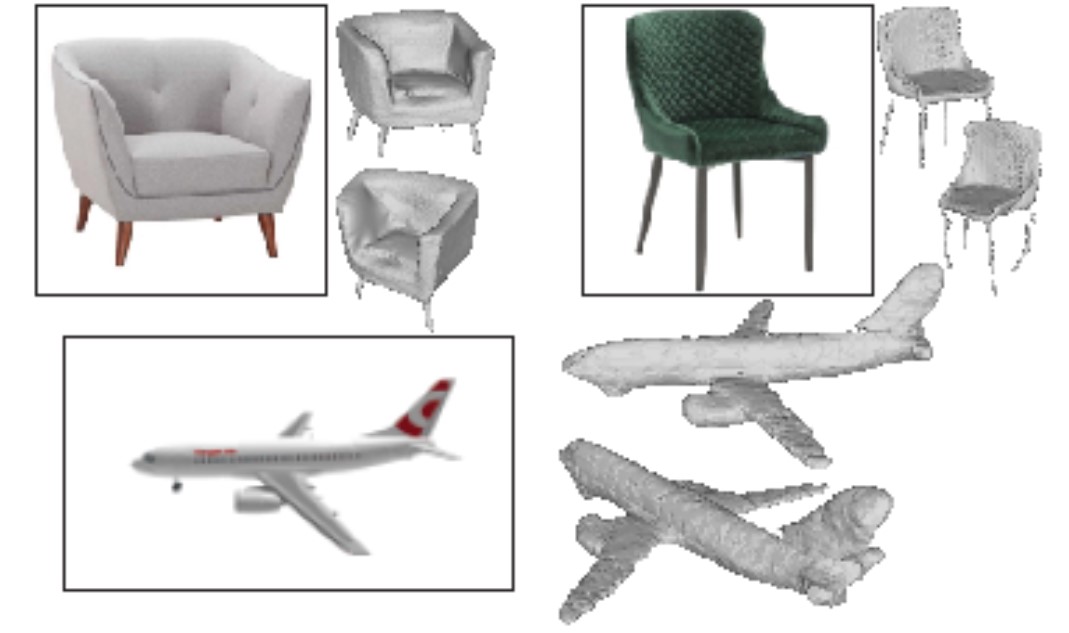}
    \caption{{\bf 3D shape from real images.} Our results (right) produced from real images (left) 
   While our results are not perfect, they provide reasonable approximation of the shown shape.}
    \label{fig:real}
    \vspace{-.3cm}
\end{figure}

\subsection{Assessing Back Map Prediction}

Figure \ref{fig:intermediate results} shows the results of our core back view from front view prediction step for several different-class examples. The depth and normal maps clearly convey that our networks can successfully predict the backside of the raw input images, replicating front features when appropriate and consistently completing entirely invisible features such as the back of the sofa or the occluded chair legs. 


\parag{Back Prediction Ablation.} To verify that the symmetry information improves back map prediction, we tested our prediction method with and without the reflected maps. 
We measure this impact on ground truth front maps of models from the test set to decouple performance of back prediction from the rest of the method.  Table~\ref{table:ablationFront2Back} shows the impact of using reflected maps on the accuracy of the back prediction for airplanes. 
We report performance in terms of average $L_1$ distance for the depth and average one minus cosine distance for the normal map; effectively this results in Eq~\ref{eq:gan}. As the numbers consistently show the use of reflection maps during prediction is important for the accuracy of resulting back maps. Figure~\ref{fig:intermediate results}, illustrates this difference on real image data. The impact of symmetry is most apparent in the normal maps, where chair legs with different normals are easily recognizable. 

\begin{figure*}
    \centering
    \includegraphics[width=\linewidth]{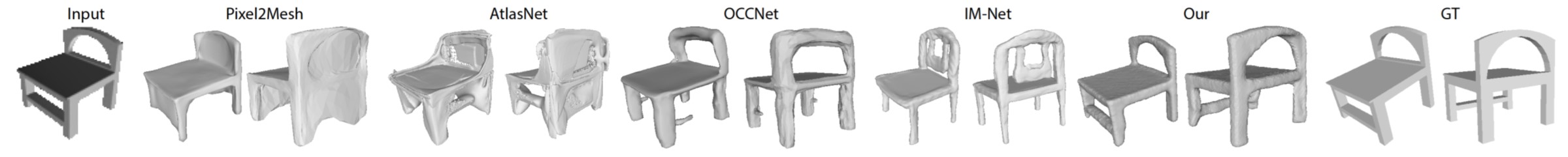}
    \includegraphics[width=\linewidth]{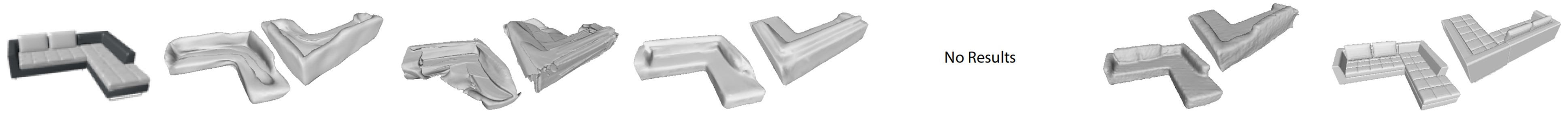}
   \includegraphics[width=\linewidth]{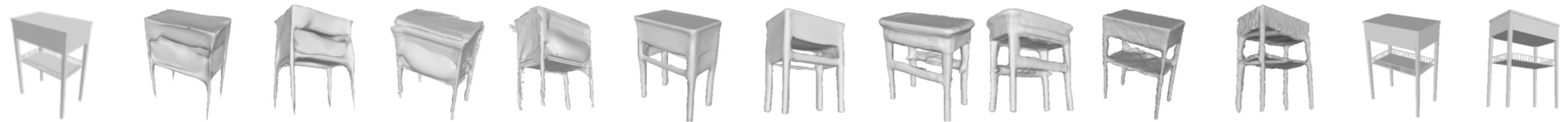}
    \caption{{\bf Qualitative comparison to state-of-the-art.} Visual comparison of our results and those produced by Pixel2Mesh \cite{wang2018pixel2mesh}, AtlasNet \cite{groueix2018}, OccNet~\cite{Mescheder19}, and IM-NET \cite{chen_cvpr19} . In all examples our results are more consistent with the input images. For some of the methods, e.g. \cite{Mescheder19,chen_cvpr19} the strong shape priors result in meshes that are close to what the network considers a reasonable object but very far away from the input image.}
        \vspace{-.3cm}
    \label{fig:comparison}
\end{figure*}

\begin{table}[h]
    \centering
    \small
{    
	\setlength{\extrarowheight}{1.5pt}
    \begin{tabular}{ l c }
        \specialrule{.1em}{.05em}{.05em} 
                             & MD   \\
        \hline
        Ours (no symmetry)     & 0.0132      \\
        Ours (with symmetry) & \textbf{0.0129}     \\
        \specialrule{.1em}{.05em}{.05em} 
    \end{tabular}
}
    \caption{\textbf{Surface ablation study} on airplane models, measuring  mesh-to-mesh distance to ground truth on surfacing outputs generated without (top) and with (bottom) reflection maps. Our outputs are closer to ground truth  and visually more symmetric (Figure~\ref{fig:ablation}).}
    \label{table:ablation}
\end{table}

\begin{table}[h]
    \centering
\resizebox{\columnwidth}{!}
{    
	\setlength{\extrarowheight}{1.5pt}
    \begin{tabular}{ l c c}
        \specialrule{.1em}{.05em}{.05em} 
                             		& depth (avg $L_1$)  & normal (avg $[1-cos(a',a)]$)\\
        \hline
        Ours (no sym)     	& 0.000593 & 0.00310    \\
        Ours (with)      		& \textbf{0.000578} & \textbf{0.00268}      \\
        \specialrule{.1em}{.05em}{.05em} 
    \end{tabular}
}

    \caption{\textbf{Front map prediction ablation:}  on airplanes. Ground truth front maps are used with (bottom row) and without (top row) reflected maps. Performance is reported in terms of average per pixel and across model $L_1$ and 1 minus cosine distance of depth and normal back predictions respectively. Clearly the use of reflected maps is shown to be beneficial. }
    \label{table:ablationFront2Back}
\end{table}

\comment{
	\begin{table}[h]
		\centering
		\setlength{\extrarowheight}{1.5pt}
		\begin{tabular}{ l c c c }
			\specialrule{.1em}{.05em}{.05em}
			Methods & chair & car & airplane\\
			\hline
			ShapeHD \cite{zhang2018shaphehd} (pre-trained) & 0.096 & 0.078 & 0.068   \\
			ShapeHD \cite{zhang2018shaphehd} (re-trained) & 0.050 & 0.035 & 0.030   \\
			Ours    & \textbf{0.037} &  \textbf{0.030}  &  \textbf{0.022}       \\
			\specialrule{.1em}{.05em}{.05em}    
		\end{tabular}
		
		\caption{{\bf Quantitative comparison to state-of-the-art.} Reconstruction errors of our method compared to ShapeHD \cite{zhang2018shaphehd} (retrained on our data). We report the chamfer distance of the reconstruction to the ground truth (lower is better). We note that retrained ShapeHD model achieves better performance than reported in original paper because our dataset, by design, is based on informative and non-accidental views of objects. Despite this, our method still produces much more accurate results than ShapeHD.}
		\label{table:1}
	\end{table}
}

\parag{3D Reconstruction Ablation.}
The final surface reconstruction uses as input a union of points from the front, reflected front, and predicted back views. To evaluate the importance of using these maps for reconstruction, we perform an ablation study on chairs and airplanes where we only feed some of the available maps into the surfacing software (Figure~\ref{fig:ablation}).  Somewhat self-evidently (Figure~\ref{fig:ablation}) accurate reconstruction is essentially impossible absent back view information.  More interestingly, our measurements (Table~\ref{table:ablation}) show that including the reflected maps in the point cloud used for final reconstruction makes a more subtle but important impact on the quality of the results.
The qualitative impact of incorporating the reflected map points is also quite significant (see Figure~\ref{fig:ablation}) - human observers expect many man-made objects to be symmetric, and reconstructions that do not respect symmetry appear far less satisfactory from a user perspective. The results incorporating symmetry are thus both better quantitatively and more visually realistic or believable. 

\begin{table*}[t]
    \tiny
    \centering
    \setlength{\extrarowheight}{1.5pt}
    \begin{tabular}{ l  |  l  | c c c c c c c c c c c c c | c}
    \multicolumn{2}{c|}{} 	   &  \multicolumn{13}{c|}{\sc Category}  \\  \cline{3-15}
    & \multicolumn{1}{c|}{\sc Methods}  & \rotatebox[origin=l]{90}{chair} & \rotatebox[origin=l]{90}{plane}  & \rotatebox[origin=l]{90}{car}   & \rotatebox[origin=l]{90}{bench}   & \rotatebox[origin=l]{90}{cabinet} 
                       & \rotatebox[origin=l]{90}{display} & \rotatebox[origin=l]{90}{lamp}  & \rotatebox[origin=l]{90}{speaker}   & \rotatebox[origin=l]{90}{rifle}   & \rotatebox[origin=l]{90}{sofa} 
                       & \rotatebox[origin=l]{90}{table} & \rotatebox[origin=l]{90}{phone}  & \rotatebox[origin=l]{90}{vessel}   & {\sc Mean} \\ \hline
    \multirow{5}{*}{\rotatebox[origin=l]{90}{Surface, Metro~\cite{Cignoni1998}}} 	
 & \scriptsize  Ours                                            & \textbf{0.013} & \textbf{0.013} & 0.013          & \textbf{0.014} & 0.014          & \textbf{0.014} & \textbf{0.019} & \textbf{0.019} & \textbf{0.012} & 0.015          & \textbf{0.012} & 0.012          & \textbf{0.016} & \textbf{0.0144} \\
& \scriptsize ONet ({ Mescheder \etal \cite{Mescheder19}})   & 0.019          & 0.016          & 0.017          & 0.017          & 0.017          & 0.022          & 0.033          & 0.036          & 0.016          & 0.019          & 0.020          & 0.021          & 0.021          & 0.0213          \\
& \scriptsize AtlasNet ({ Groueix \etal \cite{groueix2018}})      & 0.018          & 0.014          & 0.016          & 0.016          & 0.018          & 0.016          & 0.028          & 0.025          & 0.013          & 0.019          & 0.021          & 0.012          & 0.018          & 0.0181          \\
& \scriptsize Pixel2Mesh ({ Wang \etal \cite{wang2018pixel2mesh}}) & 0.016          & 0.020          & \textbf{0.011} & 0.016          & \textbf{0.012} & 0.016          & 0.021          & 0.020          & 0.014          & \textbf{0.014} & 0.014          & \textbf{0.011} & 0.021          & 0.0160          \\
& \scriptsize IM-NET ({ Chen \etal \cite{chen_cvpr19}})        &      0.023        & 0.017          & 0.018          & /              & /              & /              & /              & /              & 0.015          & /              & 0.029          & /              & /              & 0.0206          \\
\hline		                                                   
    \multirow{5}{*}{\rotatebox[origin=l]{90}{Chamfer-$L_1$~\cite{Mescheder19}}} 	
 & \scriptsize  Ours                                            & \textbf{0.021} & \textbf{0.017} & 0.019          & \textbf{0.021} & 0.023          & \textbf{0.020} & \textbf{0.023} & \textbf{0.027} & \textbf{0.015} & 0.023          & \textbf{0.019} & 0.017          & \textbf{0.022} & \textbf{0.0206} \\
& \scriptsize ONet ({ Mescheder \etal \cite{Mescheder19}})   & 0.028          & 0.023          & 0.021          & 0.022          & 0.028          & 0.031          & 0.041          & 0.047          & 0.020          & 0.025          & 0.028          & 0.028          & 0.027          & 0.0283          \\
& \scriptsize AtlasNet ({ Groueix \etal \cite{groueix2018}})      & 0.027          & 0.021          & 0.020          & 0.022          & 0.027          & 0.023          & 0.038          & 0.035          & 0.017          & 0.025          & 0.032          & 0.017          & 0.027          & 0.0254          \\
& \scriptsize Pixel2Mesh ({ Wang \etal \cite{wang2018pixel2mesh}}) & 0.022          & 0.025          & \textbf{0.016} & 0.021          & \textbf{0.019} & 0.022          & 0.028          & 0.029          & 0.018          & \textbf{0.019} & 0.022          & \textbf{0.015} & 0.028          & 0.0221          \\
& \scriptsize IM-NET ({ Chen \etal \cite{chen_cvpr19}})       & 0.035          & 0.024          & 0.021          & /              & /              & /              & /              & /              & 0.017          & /              & 0.043          & /              & /              & 0.0280          \\
                                                       		\hline


    \end{tabular}
    \caption{{\bf Comparisons against state-of-the-art.} We compare our results against  Pixel2Mesh \cite{wang2018pixel2mesh}, AtlasNet \cite{groueix2018}, OccNet~\cite{Mescheder19}, and IM-NET \cite{chen_cvpr19} measuring both mesh-to-mesh distance (MD) and $L_1$ Chamfer Distance (CD). Our method provides the best results overall for both metrics, outpeorforming the closest competitors on nine out of 13 classes.}
    \vspace{-.3cm}
    \label{table:mainresult}
\end{table*}

\subsection{Single view 3D Shape Reconstruction }

We tested our method on both 137 and 256 resolution images from the ShapeNet dataset~\cite{chang2015shapenet} as discussed above. Representative results are shown in figures~\ref{fig:teaser}, \ref{fig:methodOverview}, \ref{fig:comparison}, and  \ref{fig:real}. 
These results show that our method generates high quality results across a large number of classess, and object geometries. 

\parag{Comparisons.}
We compare our method to a range of state-of-the-art techniques, including Pixel2Mesh \cite{wang2018pixel2mesh}, AtlasNet \cite{groueix2018}, OccNet~\cite{Mescheder19}, and IM-NET \cite{chen_cvpr19} (Table~\ref{table:mainresult}). 
We use published codes and pre-trained weights from respective papers to reproduce all results.
We are unable to provide direct comparison to 3DN ~\cite{wang20193dn} because of issues with their published code\footnote{\url{https://github.com/laughtervv/3DN/issues/4}}.
Figure~\ref{fig:comparison} shows some representative comparisons. As demonstrated across multiple inputs our method consistently captures fine model details more accurately than other methods.
We note that while we use the CD metric from \cite{Mescheder19} to report the performance, we are recomputing all the numbers. Both \cite{Mescheder19} and Chen \etal ~\cite{chen_cvpr19} rely on watertight training models, that only approximate ground truth,  and report Chanfer distance wrt to these proxies,  while we seek to measure distances wrt to the real ground truth. 
Table~\ref{table:mainresult} verifies quantitatively what the preceding qualitative results have shown: Our reconstructions are much closer to the ground truth shapes than those of other learning-based shape reconstruction methods, reducing both MD and CD by as much as 19\% (on cars) and 18\% (on lamps) respectively. On average across the 13 categories we are 12.5\% better than the closes competing method, which turns out to be \cite{wang2018pixel2mesh}. Overall our method is more accurate on 9 out of 13 categories; we perform worse on {\em cars}, {\em cabinets}, {\em phones} and {\em sofa} where we are only marginally worse than the best method.


\parag{Application to real images.}
Figure \ref{fig:real} shows that, although trained on synthetic renderings, our method yields realistic reconstructions on real images when provided with a segmentation mask or simple background.

\section{Conclusion}

We presented a novel single view 3D reconstruction method anchored around prediction of back view maps from front view information and demonstrated  it to improve on the state of the art. Key to the success of our method is a combination of advanced learning approaches with geometry priors that motivate our algorithmic choices.  
Notably, our prediction of 2.5D front maps from images can be potentially substituted by other sources of depth and normal maps, such as depth scanners or sketch processing systems, enabling direct reconstruction of complete 3D objects from such data.



\paragraph{Acknowlegements:}  This work was funded in part by the Vector Institute for AI, Canada CIFAR AI Chair, NSERC Canada Research Chair (CRC), Compute Canada RRG, and NSERC Discovery Grants.

{\small
\bibliographystyle{cvpr}
\bibliography{main}
}

\newpage

\section*{Appendices}
\appendix

In this document, we supply additional evaluation, training, and implementation
details, and provide a more detailed ablation study.

\section{Additional qualitative results} 
We included only few qualitative experiments in the main paper due to space limitation. In Figure~\ref{fig:comparison} we provide additional experiments that illustrate performance on a broader set of classes and compare our results against state-of-the-art alternatives. In all cases we are able to recover more faithful topology and geometry of objects, including local high-fidelity detail.  Our results consistently more accurately reflect the ground-truth geometry.  Among all methods compared against we come closest in approximating the geometry of the lamps (rows 1, 6), and capturing the details of the boats (rows 4,5) and the airplane (row 7). Contrary to Pixel2Mesh \cite{wang2018pixel2mesh} we correctly recover the topology of the input models, accurately capturing handles on the display (row 2) and bed (row 3) and correctly recovering the chair back (rows 8, 9)  and table (last row) leg connectivity.  Contrary to methods such as OccNet~\cite{Mescheder19} and IM-NET \cite{chen_cvpr19}  we do not hallucinate non-existent details (see car, row 10, gun row 11, or table, last row). 

\begin{figure*}[h]
	\centering
	\includegraphics[width=\linewidth]{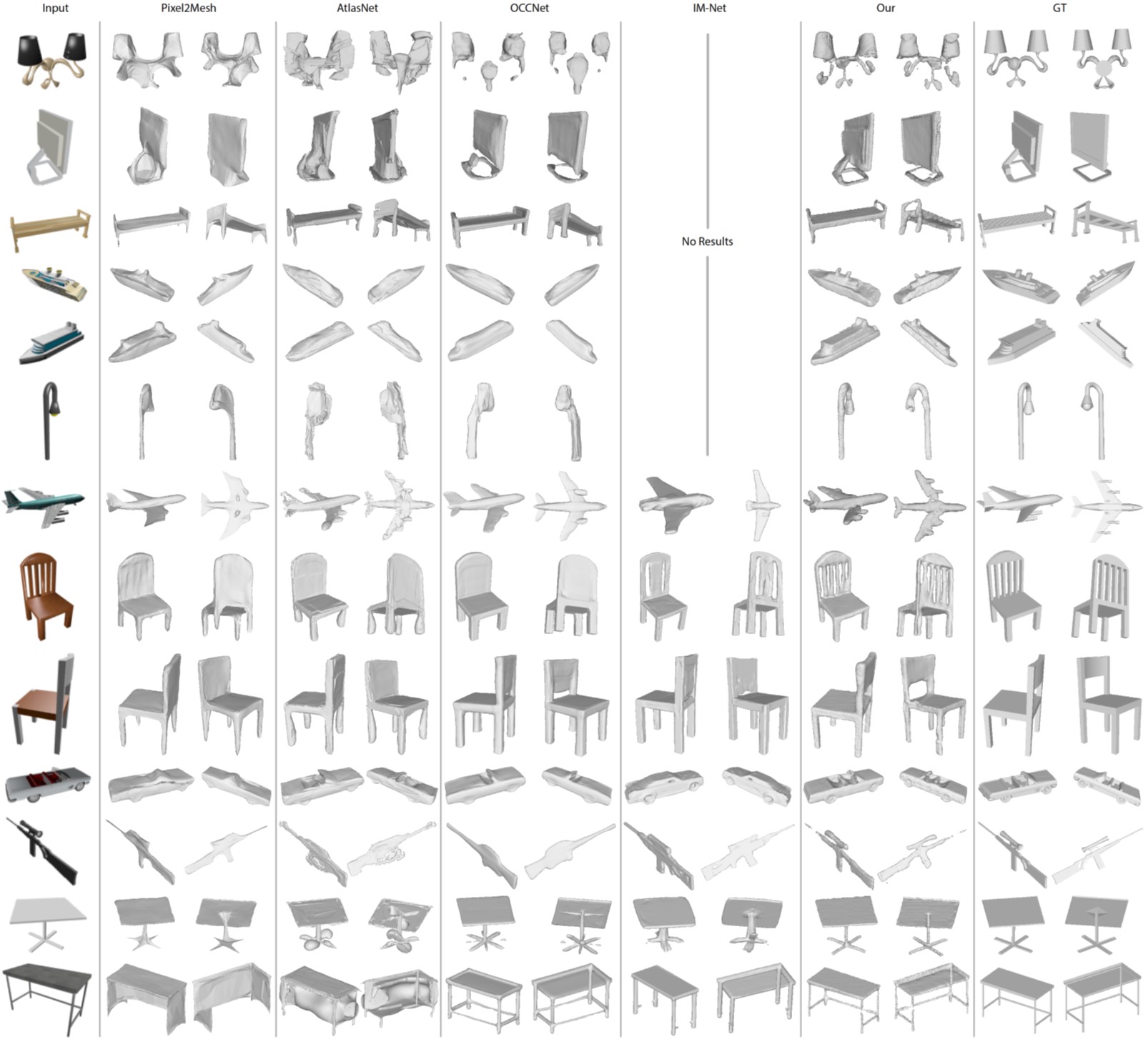}

	\caption{{\bf Qualitative comparison to state-of-the-art.} Visual comparison of our results and those produced by Pixel2Mesh \cite{wang2018pixel2mesh}, AtlasNet \cite{groueix2018}, OccNet~\cite{Mescheder19}, and IM-NET \cite{chen_cvpr19}. Our results can recover the shape of the input well and capture much more details than others, without halucinating non-existent details. Note that as IM-NET \cite{chen_cvpr19} only provides pre-trained models on five classes, we cannot compare with them on other classes such as  lamps, displays, etc.}
	\vspace{-.3cm}
	\label{fig:comparison}
\end{figure*}

\section{Additional quantitative evaluation}


Due to lack of space, in the main paper we only report Chamfer $L_1$ distance (CD) measured using the implementation in \cite{Mescheder19} and surface-to-surface distance (MD), as computed by Metro \cite{Cignoni1998}, as our error metrics. Here we include additional evaluation, with respect to state-of-the-art, based on the Normal Consistency metric introduced by ~\cite{Mescheder19}. Results  are reported in Table~\ref{table:normal}. Since we measure consistency, higher numbers indicate better performance, with 1 being the optimum.  Our proposed method works better on $8$ out of $13$ categories and overall. This provides additional evidence that we improve on the previous state-of-the-art performance.   Since we measure accuracy directly against unprocessed ShapeNet models, most of which are not watertight, we do not provide IoU measurements, since those are only valid for comparing watertight meshes.

\vspace{0.07in}
\noindent
{\bf Additional information on experiments.}
When comparing our results to other methods we encountered two challenges. First, a number of methods, \eg OccNet~\cite{Mescheder19} and IM-NET \cite{chen_cvpr19} that use watertight proxies of ShapeNet models for training, measure result accuracy against these proxies, instead of the originating ShapeNet models. We seek to measure accuracy against the original ShapeNet models. Second, while we followed the original ShapeNet train/test/val split to evaluate our models, other methods \eg Pixel2Mesh \cite{wang2018pixel2mesh}, AtlasNet \cite{groueix2018} use the split provided by ~\cite{choy20163d}.
To resolve both issues we use pre-trained models provided by the authors of the relevant papers and test them on the intersection of the two test sets. 
This intersection test set contains 1685 models: 197 airplanes, 83 benches, 62 cabinets,
149 cars, 324 chairs, 48 displays, 101 lamps, 63 speakers, 75 rifles, 132 sofas, 339 tables,
46 cellphones, and 66 boats.
We then measure all reported metrics on these results, comparing them against the original ground truth ShapeNet \cite{chang2015shapenet} models. 


Each method we compare to uses a different scaling convention. We apply the transformations used by those methods to transform their outputs accordingly to match ground truth models. For AtlasNet~\cite{groueix2018}, OccNet~\cite{Mescheder19} and Pixel2Mesh~\cite{wang2018pixel2mesh}, we do the exact transformations they described in their papers and supplementary materials. For IM-NET~\cite{chen_cvpr19} we follow the transformations used in the voxelization process they utilized \cite{hane2017hierarchical}. 


%
%
%
%



\begin{table*}[th]
	\scriptsize
	\centering
	\setlength{\extrarowheight}{1.5pt}
	\begin{tabular}{ l  |  l  | c c c c c c c c c c c c c | c}
		\multicolumn{2}{c|}{} 	   &  \multicolumn{13}{c|}{\sc Category}  \\  \cline{3-15}
		& \multicolumn{1}{c|}{\sc Methods}  & \rotatebox[origin=l]{90}{chair} & \rotatebox[origin=l]{90}{plane}  & \rotatebox[origin=l]{90}{car}   & \rotatebox[origin=l]{90}{bench}   & \rotatebox[origin=l]{90}{cabinet} 
		& \rotatebox[origin=l]{90}{display} & \rotatebox[origin=l]{90}{lamp}  & \rotatebox[origin=l]{90}{speaker}   & \rotatebox[origin=l]{90}{rifle}   & \rotatebox[origin=l]{90}{sofa} 
		& \rotatebox[origin=l]{90}{table} & \rotatebox[origin=l]{90}{phone}  & \rotatebox[origin=l]{90}{vessel}   & {\sc Mean} \\ \hline
		                                                   
		& \scriptsize  Ours                                            & \textbf{0.771}& \textbf{0.759}  & 0.734           & 
		0.674 		   & 
		0.763           & 
		0.821 		   & \textbf{0.734}  & \textbf{0.780}  & \textbf{0.672}  & 
		\textbf{0.781}  & \textbf{0.789}  & 
		0.856           & \textbf{0.733}  & \textbf{0.759} \\
		& \scriptsize ONet ({ Mescheder \etal \cite{Mescheder19}})   & 0.741          & 0.719          & \textbf{0.759}          & 0.668          & 0.759          & 0.771          & 0.707          & 0.729          & 0.590          & 0.757          & 0.725          & 0.824          & 0.667          & 0.724          \\
		& \scriptsize AtlasNet ({ Groueix \etal \cite{groueix2018}})      & 0.701          & 0.682          & 0.725          & 0.627          & 0.725          & 0.772          & 0.641          & 0.738          & 0.549          & 0.734          & 0.671          & 0.839          & 0.614          & 0.694          \\
		& \scriptsize Pixel2Mesh ({ Wang \etal \cite{wang2018pixel2mesh}}) & 0.741        & 0.747          & 
		0.698          & 		\textbf{0.685} & \textbf{0.782} & \textbf{0.837} & 0.705          & 0.767          & 0.636          & 
		0.780 		   & 0.762          & \textbf{0.881} & 0.681          & 0.746          \\
		& \scriptsize IM-NET ({ Chen \etal \cite{chen_cvpr19}})        &0.686          & 0.687          & 0.680          & /              & /              & /              & /              & /              & 0.579          & /              & 0.657          & /              & /              & 0.658          \\
		\hline

	\end{tabular}
	\caption{{\bf Normal Consistency~\cite{Mescheder19} comparisons against state-of-the-art.} We compare our results against  Pixel2Mesh \cite{wang2018pixel2mesh}, AtlasNet \cite{groueix2018}, OccNet~\cite{Mescheder19}, and IM-NET \cite{chen_cvpr19} measuring Normal Consistency (larger value better). Our method provides the best results overall and outperforms the closest competitors on 8 out of 13 classes.}
	\vspace{-.3cm}
	\label{table:normal}
\end{table*}

\section{Scalability to High Resolution}
\begin{figure}[h]
    \centering
    \includegraphics[width=\linewidth]{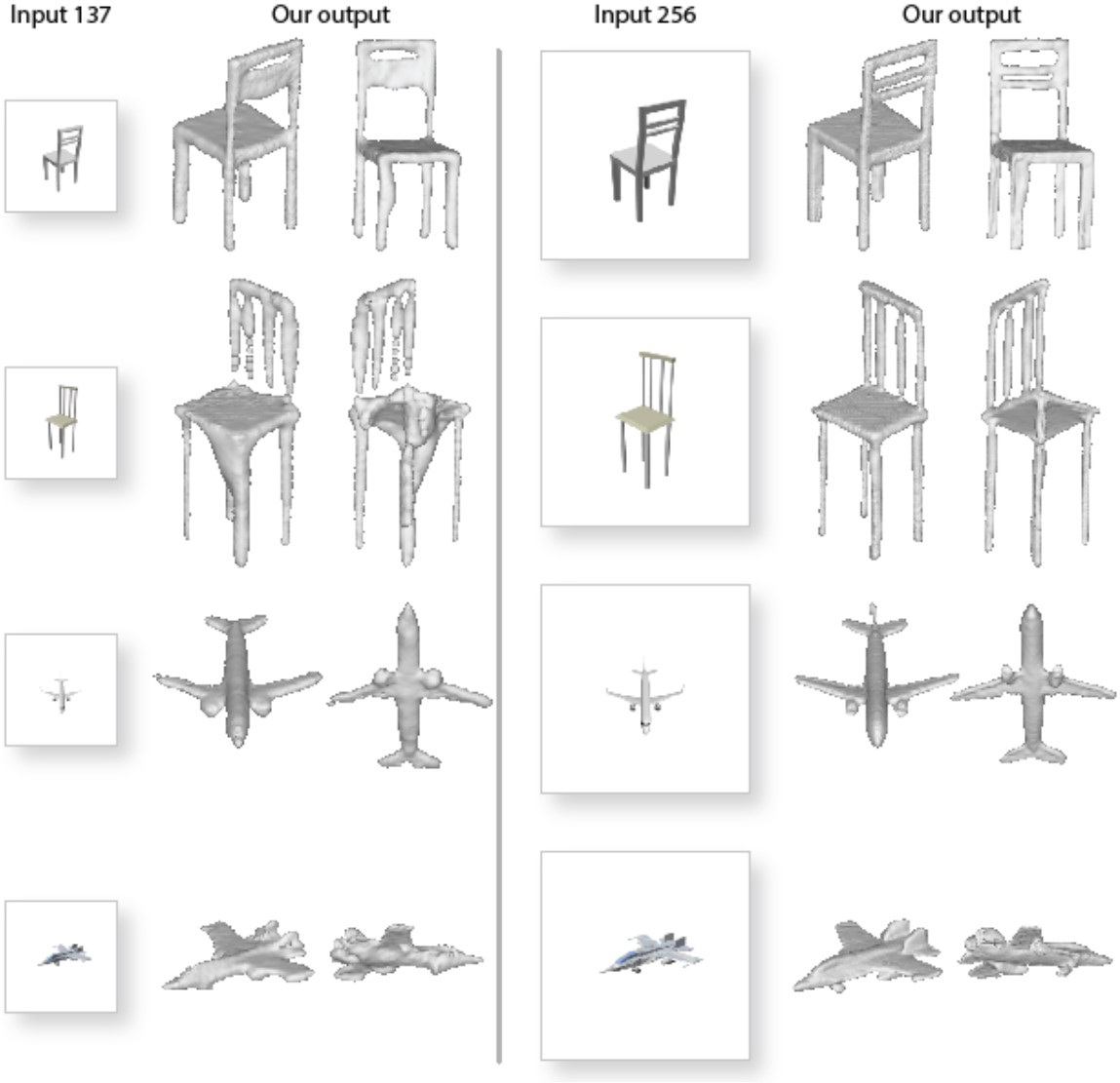}
    \caption{{\bf $137\times137$ vs $256 \times 256$ resolution results.} (left) low resolution images and results. (right) similar view and render style   $256 \times 256$ images  and corresponding results. 
  The quality of our reconstruction predictably improves with increase in image resolution.}
    \label{fig:large}
\end{figure}

Our Front2Back map-based formulation can easily scale to operate at a higher resolutions.  To do so, we simply need to train (i) image to front map and (ii) front2back modules to operate at the higher resolution (\ie, take higher resolution image as input and produce, equivalent, higher resolution depth and normal maps as outputs). The rest of our framework can remain exactly as is. 
To illustrate this beneficial capability, we report additional experiments we didn't have space to include in the main paper. We report performance at 256$\times$256 resolution on the {\tt chair} and {\tt plane} classes. Results are qualitatively illustrated in Figure~\ref{fig:large} and quantitatively  analyzed in Table \ref{table:highres}. Higher resolution maps allow recovery of finer geometric detail observed in the outputs. 
For example, in our higher-resolution results, the legs on the chairs appear much cleaner and geometrically more accurate (row 2) and planes have much more decernable engines and other fine detail (row 3 and 4).
This is quantified in Table \ref{table:highres}, which shows significantly improved reconstruction scores (approximately 15\% improvement on average) as compared to our main 137$\times$137 resolution model.

\begin{table}[!h]
	\centering
	\setlength{\extrarowheight}{1.5pt}
	\begin{tabular}{ c  |  c  | c c }
		
		& \multicolumn{1}{c|}{\sc Methods}  & chair & plane  \\ \hline
		\multirow{2}{*}{MD} 	
		& \scriptsize  Ours - 137  & 0.013 & 	0.013  \\
		& \scriptsize Ours - 256   & \bf{0.011}          & \bf{0.011}                \\

		\hline		                                                   
		\multirow{2}{*}{CD} 	
		& \scriptsize  Ours - 137 & 0.021 & 	0.017  \\
		& \scriptsize Ours - 256   &   \bf{0.018}      & \bf{0.015}               \\

		\hline

	\end{tabular}
	\caption{Quantitative results on high resolution (256$\times$256) inputs versus low resolution (137$\times$137) inputs. The performance is increased by 10\%-20\%.}
	\vspace{-.3cm}
	\label{table:highres}
\end{table}

\section{Implementation Details}
\label{sec:implementation}
In this section, we add more details for our implementation which is not listed in the paper. 

\parag{Symmetry plane detection.} 
We represent the symmetry planes as 3-dimensional vectors $(\phi, \theta, d)$, where $\phi, \theta$ are the polar angles of the plane's orientation, and $d$ is the distance of the plane to the origin. We use mean-shift clustering with the Epanechnikov kernel to generate an initial symmetry plane guess. 



\parag{Neural network architecture.}
We train pix2pix~\cite{isola2017image} with the resnet-9 block architecture to predict the back view oriented depth maps. Our input images are 256$\times$256 pixels and have eight channels (three channels for the normals, one channel for the depth; both for front view and reflected view). In case of 137$\times$137 model we pad to 256; for the high-resolution model we directly use 256$\times$256 inputs. For the front normal and depth map prediction we train separate networks for 137$\times$137 and 256$\times$256 resolutions. 

\comment{
	\alla{in correct terminology note that we minimize the sum of the loss functions across the three produced maps}
	For the 2D to 2.5D step we trained the 2.5D estimation network of \cite{wu2018learning}. The network of  \cite{wu2018learning} is designed for noisy, real-life images and purposefully adds noise to input data to mimic real-life artifacts; since similar to most recent single view reconstruction methods our training image set is synthetic, for a fair comparison we disabled this feature.  We subsequently mask the depth and normal maps using the silhouette map, denoting all pixels outside it as background. 
}

\parag{Training scheme.}

We train separate models for the image to 2.5D map representation and the back2front reconstruction. Both networks are trained independently on ground truth data. The learning rate is set to $2e-4$ and decays by 0.5 after $20$ epochs; we train for a total of $40$ epochs. Our training data set contains independent classes such as car, airplane, \etc; we train separate models for each class. This is consistent with other papers in the area.

\comment{
For the final surface reconstruction, we first generate the oriented point cloud by combining the front, reflected, and predicted back maps. Notably for each pixel within the predicted object silhouette we can have either two or three pairs of depth+normal values, one from each map (the reflected map typically covers only a portion of  these pixels, Figure~\ref{fig:reflected}). Basic geometric analysis states that given front and back map depth values for the same pixel a correct front map depth value should always be closer to the observer than the back one. Since back maps are predicted from the front maps, we expect the front prediction to be more accurate.  We consequently ignore back map depth+normal pairs that violate this prior when assembling the point-cloud. We apply a similar logic when assessing consistency between reflected maps and back or front ones. Specifically, since reflected maps are generated from the front ones using well defined priors, we expect front maps to be more reliable than reflected ones, and reflected to be more reliable than back. Consequently given front and reflected map depth values for the same pixel, we ignore reflected map values whenever the reflected depth is closer to the observer than the front one, and given back and reflected map depth values for the same pixel, we ignore back map values whenever the back depth is farther from the observer than the reflected one. 

We subsequently remove outliers from this cloud using the following observation - while we expect our data to contain depth discontinuities we expect prominent features (which have a distinctly different depth than that of their surroundings) to have a front (or back) facing  area bigger than a single pixel. Thus we filter out points if the minimal depth differential between their depth and that of their four immediate neighbors in image space is above a fixed threshold (1.5\% of the image depth range in our implementation \alla{check again the number}).  After that, we use the source code provided by the authors of Poisson surface reconstruction~\cite{kazhdan2013screened} to obtain shape reconstruction result. To allow reconstruction of the details visible in the $256 \times 256$ input images while generating smooth mesh, we set the octree depth of the internal volumetric representation to $7$. \alla{any other parameters?}
}

\section{Limitations.}

Our method performs well on typical views of everyday objects, where the front and back views (combined with the reflection of the front) describe a big portion of the shape.
Predictably, its performance declines on accidental views  where the union of these information sources is insufficient to describe the surface (see Figure~\ref{fig:limitations} a). A possible solution for such scenarios is to perform surface completion, 
\eg using the method of \cite{Mescheder19};  or to leverage our existing algorithm rendering the incomplete output from a different direction (ideally one maximally opposite to the incomplete data); repeating the back prediction step, and performing reconstruction using a union of previous and new maps. 
Our framework does not hallucinate geometry not evident from the images, thus given fuzzy, ambiguous images it is not able to produce meaningful reconstructions (Figure~\ref{fig:limitations} b). 
Symmetry computation on raw images or predicted front maps is a challenging problem; incorrect symmetry plane  estimation can potentially lead to  reconstruction failures (see Figure~\ref{fig:limitations} c). 
We minimize the likelihood of such failures by using very conservative symmetry detection criteria; more robust detection of symmetry planes on image inputs is an interesting future research topic. 


\begin{figure}[!h]
	\centering
	\includegraphics[width=1\linewidth]{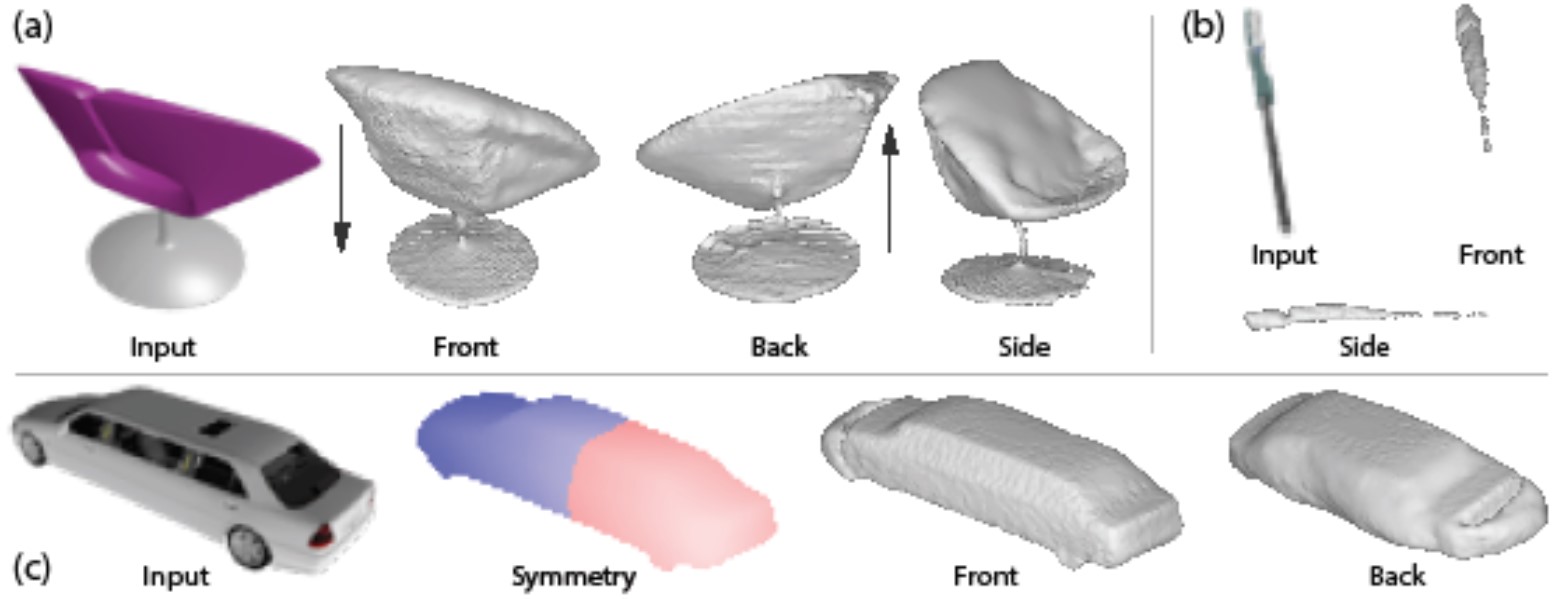}
	
	\label{fig:limitations}
	\caption{{\bf Limitations.}  (a) Our method's performance declines on accidental view images, such as these, where important details are not discernible from front or predicted back views. (b) Our method cannot recover fuzzy details from ambiguous low-resolution inputs.  (c) A wrongly detected reflective symmetry (here left/right instead of front/back) can lead to reconstruction artifacts.}
	\vspace{-.3cm}
	\label{fig:limitations}
\end{figure}

\end{document}